\documentclass[runningheads]{llncs}

\usepackage{eccv}

\usepackage{eccvabbrv}

\usepackage{graphicx}
\usepackage{booktabs}
\usepackage{tabularx}
\usepackage{multirow}
\usepackage{rotating}
\usepackage{comment}
\usepackage{tikz}
\usepackage{pgfplots}
\usepackage{pgfplotstable}
\pgfplotsset{compat=1.18}
\usepackage{dirtree} %
\usetikzlibrary{shapes,arrows.meta,positioning,fit}

\usepackage[accsupp]{axessibility}  %

\usepackage{hyperref}

\usepackage{orcidlink}

\usepackage{siunitx}

\PassOptionsToPackage{nameinlink}{cleveref}

\newrobustcmd{\todo}[1]{\textcolor{orange}{#1}}

\begin{document}

\title{Empowering Autonomous Shuttles with Next-Generation Infrastructure}
\author{Sven Ochs\inst{1} \and %
        Melih Yazgan \inst{1}\orcidlink{0009-0007-5773-1751} \and %
        Rupert Polley \inst{1} \and %
        Albert Schotschneider \inst{1} \orcidlink{0000-0002-6376-1793} \and %
        Stefan Orf \inst{1} \and %
        Marc Uecker \inst{1}\orcidlink{0000-0003-2489-5841} \and %
        Maximilian Zipfl \inst{1,2} \and %
        Julian Burger \inst{1} \and \\ %
        Abhishek Vivekanandan \inst{1} \and %
        Jennifer Amritzer \inst{1} \and \\ %
        Marc René Zofka\inst{1}\orcidlink{1111-2222-3333-4444} \and %
        J. Marius Z\"ollner\inst{1,2}\orcidlink{0000-0001-6190-7202}}

\authorrunning{S.~Ochs et al.}
\institute{FZI Research Center for Information Technology (FZI),
Karlsruhe, Germany\\\email{\{last name\}@fzi.de} \and
Karlsruhe Institute of Technology (KIT), 
Karlsruhe, Germany\\
\email{zoellner@kit.de}}

\maketitle

\begin{abstract}
As cities strive to address urban mobility challenges, combining autonomous transportation technologies with intelligent infrastructure presents an opportunity to transform how people move within urban environments. Autonomous shuttles are particularly suited for adaptive and responsive public transport for the first and last mile, connecting with smart infrastructure to enhance urban transit. This paper presents the concept, implementation, and evaluation of a proof-of-concept deployment of an autonomous shuttle integrated with smart infrastructure at a public fair. The infrastructure includes two perception-equipped bus stops and a connected pedestrian intersection, all linked through a central communication and control hub. Our key contributions include the development of a comprehensive system architecture for "smart" bus stops, the integration of multiple urban locations into a cohesive smart transport ecosystem, and the creation of adaptive shuttle behavior for automated driving. Additionally, we publish an open source dataset and a Vehicle-to-X (V2X) driver to support further research. Finally, we offer an outlook on future research directions and potential expansions of the demonstrated technologies and concepts.

\end{abstract}

\section{INTRODUCTION}

\subsection{Motivation}
\begin{figure}[ht]
    \centering
    \includegraphics[width=\linewidth]{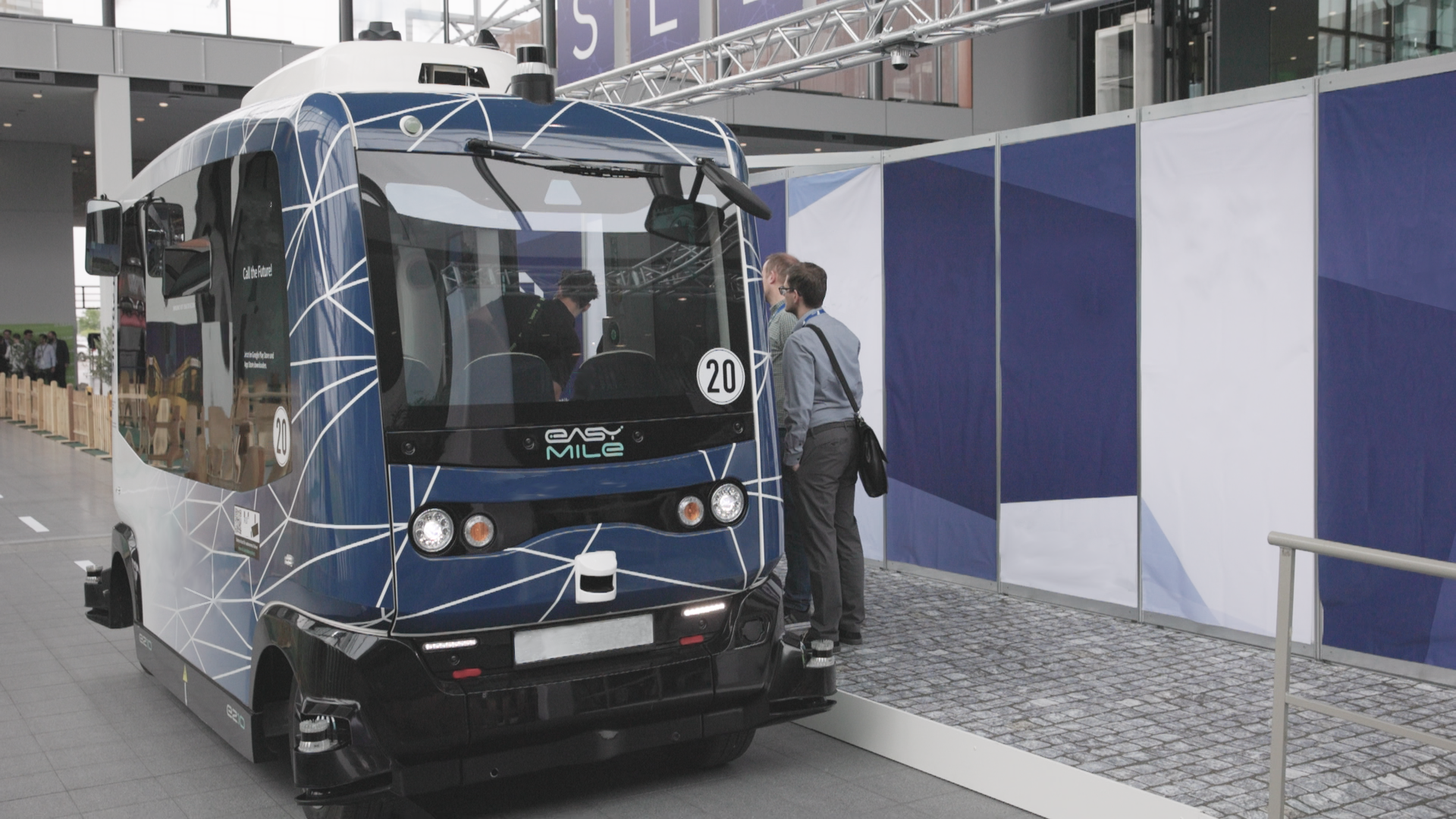}
    \caption{Smart next generation bus stop and automated shuttle. The smart next generation bus stop is equipped with a surround view LiDAR and an AI-based detection pipeline to create and distribute a local environment model.}
    \label{fig:introduction:unique_key_selling_pt_image}
\end{figure}

\begin{figure}[ht]
    \centering
    \hfill%
    \begin{subfigure}[t]{0.32\linewidth}%
        \includegraphics[width=\textwidth]{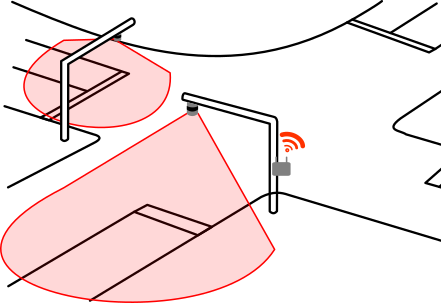}%
        \caption{Roadside infrastructure at intersections}%
        \label{fig:traffic_location_examples:first}%
    \end{subfigure}%
    \hfill%
    \begin{subfigure}[t]{0.32\linewidth}%
        \includegraphics[width=\textwidth]{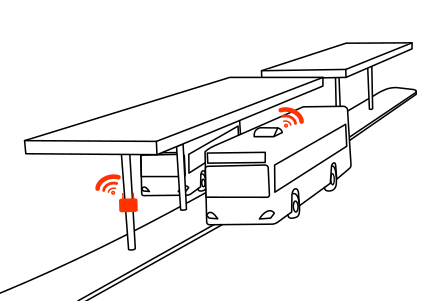}%
        \caption{Smart stops and stations for public transport}%
        \label{fig:traffic_location_examples:second}%
    \end{subfigure}
    \hfill%
    \hspace{0pt} %
    \caption{Potential traffic scenarios, that are suitable for smart infrastructure.}
    \label{fig:traffic_location_examples}
\end{figure}

Automated shuttles in public transport must meet specific requirements for operation in public traffic. Bus stops and waiting areas are vital interaction points for various Vulnerable Road Users (VRUs), including pedestrians, cyclists, vehicles, and other traffic participants. These areas serve as hubs for public transport services, providing passengers with information, ticketing, and operational services.
Traditionally, bus drivers manage numerous tasks, such as driving, adhering to schedules, and handling ticketing. Additionally, drivers perform essential safety functions at bus stops by observing and predicting crowd movements to approach the stop safely.
As autonomous and connected shuttles become more prevalent, the reduction in onboard operators necessitates reconsidering the roles traditionally performed by bus drivers. Connected infrastructure and sensor technology can significantly enhance safety in waiting areas by monitoring crowd movements. Intelligent infrastructure extends the shuttle’s field of view through additional sensors and communication capabilities, improving safety at pedestrian crossings. Signalized traffic flow can be actively managed and preemptively controlled, increasing the efficiency of automated services.
Real-world laboratories equipped with roadside infrastructure and prototype sensor setups are ideal for testing Cooperative Intelligent Transport Systems, algorithms, and AI in a realistic yet controlled environment~\cite{ochs_last_2024}. The most commonly investigated scenarios are depicted in \Cref{fig:traffic_location_examples}. These laboratories allow for evaluating communication protocols between vehicles and infrastructure using advanced sensors like cameras and LiDAR to monitor traffic flow, detect pedestrians, and assess real-time traffic conditions. This setup is crucial for developing algorithms that enhance safety, reduce congestion, and improve overall traffic efficiency. Additionally, smart stations in real-world laboratories facilitate the exploration of innovative technologies to enhance commuter experiences, optimize public transport operations, and improve safety.
\subsection{Contribution}
This paper presents the concept, implementation, and evaluation of a prototypical smart bus stop designed to support the automated driving functions of public transport shuttles. Our key contributions are as follows:
\begin{enumerate}
    \item \textbf{Concept for Smart Bus Stops:} Developed a system architecture to facilitate automated transport.
    \item \textbf{Integration of Urban Locations:} Combined smart traffic lights and bus stops into a cohesive small-scale smart transport ecosystem.
    \item \textbf{Adaptive Shuttle Behavior:} Enabled shuttles to utilize environmental models from both smart bus stops and traffic lights for improved automated driving.
\end{enumerate}
Additionally, this paper is accompanied by an open dataset (\url{https://url.fzi.de/v2x_dataset}) and an open-source Vehicle-to-X (V2X) driver (\url{https://url.fzi.de/cohda_mk6_driver}) to support and facilitate further research and development.%

\section{RELATED WORK}
\label{sec:related_work}
The concept of smart cities, leveraging digital technology to enhance urban services while reducing costs and resource consumption, has gained significant attention. Key components include intelligent infrastructure, smart transportation systems, and technology in urban planning. AI technology sets smart city systems apart from traditional infrastructure by enabling communication between components \cite{herath_adoption_2022}. Smart Infrastructure refers to integrating digital technology with physical infrastructure to monitor, control, and optimize the operation of public utilities and services such as traffic flow \cite{kasznar_multiple_2021}. An instance of a flexible and scalable setup for smart infrastructure is realized in Test Area Autonomous Driving, Baden-Württemberg \cite{fleck_towards_2019}. The proposed system handles traffic light states and road topology information and captures data on traffic participants through camera-based object detection and tracking \cite{fleck_robust_2020}.

Smart Transportation is crucial for smart cities, enhancing the safety and efficiency of public transportation through autonomous vehicles and intelligent systems. Smart infrastructure combines sporadic traffic information from specific operational design domains to create energy-efficient transportation systems\cite{wang_is_2021, nguyen_demand_2020}. The role of V2X in this context, but in another domain, is exemplified by research on automated parking systems \cite{schorner_park_2021}. This work introduces a parking management system that optimally assigns vehicles to parking spaces, considering different automation levels. V2X communication is integral, influencing both low-level vehicle path planning and high-level parking spot assignment. Messages such as Cooperative Awareness Messages (CAM), Collective Perception Messages (CPM), and Maneuver Coordination Messages (MCM) coordinate transportation system components, especially in areas with unreliable GPS accuracy. Information dissemination between infrastructure components and vehicle systems occurs via a V2X communication layer defined by messaging standards, requiring high-frequency, low-latency communication. 6G, utilizing higher frequency radio bands, offers much wider bandwidth than 5G with sub-millisecond latency. This enables a unified, interconnected network where real-time data sharing is critical, making the system robust and dependable. Continued advancements in communication technologies like 6G can accelerate and support further research and development in smart infrastructure \cite{murroni_6genabling_2023}.

Additionally, \cite{yazgan_shuttle2x_nodate} has shown the practical utilization and operation of V2X to enhance passenger transport through a functional automated shuttle. The operational safety of the highly automated driving function is enhanced by additional information from Roadside Units (RSU) via V2X. For instance, CPM messages inform the shuttle about occluded obstacles and ones outside its field of view. Furthermore, automated shuttles are extended for last-mile delivery by integrating smart cargo cages, as demonstrated by Ochs et al. \cite{ochs_last_2024}.

In the realm of smart transportation, smart bus stops are an emerging trend. Equipped with sensors and smart technologies, these bus stops enhance the commuter experience. A key component of smart bus systems is obtaining real-time passenger flow information, allowing networks to optimize bus deployment during peak times. This is demonstrated by \cite{zhang_design_2021}, where a genetic algorithm estimates passenger flow using vision-based sensors. While smart bus systems exemplify advancements in urban transportation, the development of V2X communication technologies further underscores the necessity of open datasets for enhancing smart city infrastructure.

Open datasets are crucial for advancing V2X communication in smart cities, offering essential data for developing, testing, and benchmarking urban mobility algorithms. Yazgan et al.~\cite{yazgan_collaborative_2024} emphasize the importance of V2X datasets for creating advanced V2X applications such as collaborative perception and cooperative maneuver planning. Despite the availability of simulation datasets \cite{li_v2x-sim_2022, xu_v2x-vit_2022, xu_opv2v_2022}, real-world complexities are better captured by open-sourced datasets \cite{yu_dair-v2x_2022, yu_v2x-seq_2023, zimmer_tumtraf_2024}. These datasets contain sensor data but lack European Telecommunications Standards Institute (ETSI) message implementations. The authors from \cite{mavromatis_operating_2019} present a dataset documenting network interactions between two V2X Onboard Units (OBUs) and several RSUs along the Flourish test track in Bristol, UK. This dataset, collected over eight two-hour experiments, logs CAMs every 10 milliseconds, focusing on GPS data.
In contrast, \cite{bmeautomateddrive_bmeautomateddrivev2xdatasets_nodate} introduces a multi-modal dataset from over 1800 vehicles and multiple RSUs in real-world public road traffic. It includes various different V2X messages, such as CAMs, Decentralized Environmental Notification Messages~(DENMs), Signal Phase and Timing Extended Messages~(SPATEMs), and Map Topology Extended Messages~(MAPEMs), provided in JSON format.
Meanwhile, \cite{kueppers_v2aix_2024} presents the V2AIX dataset, a multi-modal collection of ETSI messages from real-world traffic scenarios. This dataset includes CAMs, DENMs, SPATEMs, and MAPEMs, gathered from both mobile and stationary setups, designed for integration with the Robot Operating System~(ROS).

To address the gap in current datasets, we developed a small-scale indoor dataset that includes CPMs. Indoor datasets present unique challenges, such as precise time synchronization between various sensors and communication devices to ensure the integrity and accuracy of transmitted messages. Indoor positioning is particularly challenging due to unreliable GPS signals, and precise maneuvers in limited spaces increase demands on navigation and control.  By leveraging these datasets, it is possible to improve the integration of digital technologies with physical infrastructure, enhancing urban services, reducing resource consumption, and improving the overall quality of life in cities.

\section{CONCEPT}
\label{sec:method}

\begin{figure}[htb]
    \centering
    \includegraphics[width=\textwidth]{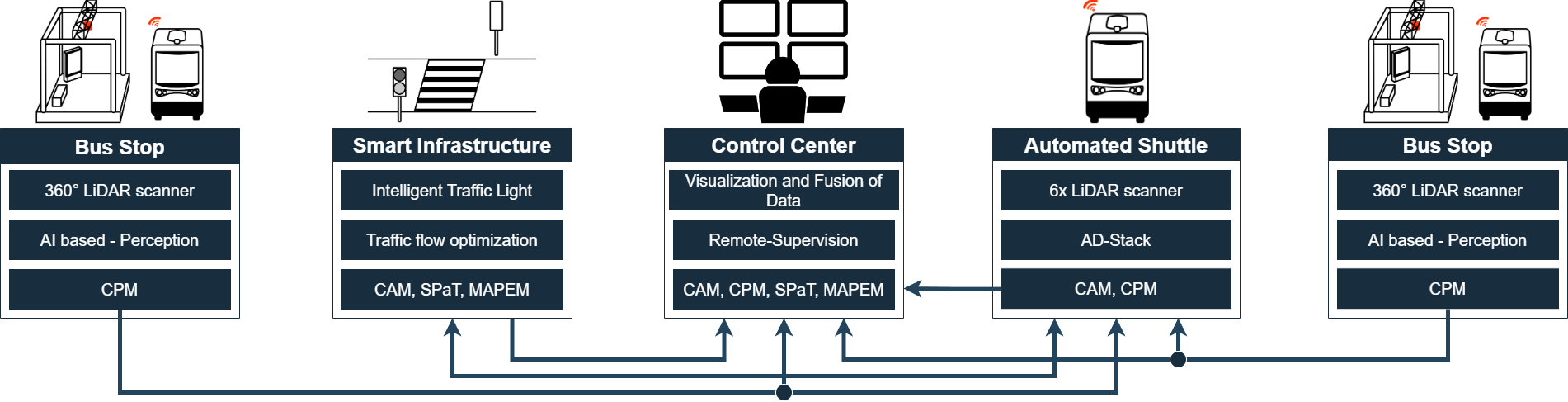}
    \caption{Overview of our setup: showcasing all components and their unique capabilities, with highlighted communication pathways.}
    \label{fig:method:architecture}
\end{figure}
The future of mobility includes robo-taxis, on-demand last-mile shuttles, and automated bus lines. This paper combines the benefits of automated buses and last-mile shuttles with pickup points at train/bus stops, schools, and tourist attractions. These stops must support shuttle driving functions for safe operation, as shuttles can only detect people within their sensor range. Detecting and predicting group behavior requires smart infrastructure to extend the Field of View (FOV) of the automated shuttle, utilizing the Intelligent Transport Systems (ITS)-G5 standards.
The overall architecture is visualized in \Cref{fig:method:architecture}. The primary elements of this demonstration include:

\textbf{V2X Communication:} The shuttle sends CAM, and the infrastructure sends SPATEM and MAPEM messages to manage traffic light states and road topology information. Bus stops provide passenger information to the shuttle via CPM.

\textbf{Smart Infrastructure:} This is a pedestrian crossing designed to ensure pedestrian safety while coordinating with the shuttle's movements. It includes screens that display intuitive signals to pedestrians and communicate traffic light status to the shuttle using V2X technology.

\textbf{Smart Bus Stops:} Equipped with advanced LiDAR sensors and AI-based detection systems, these bus stops create and share a local environment model. They communicate via V2X with the shuttle to provide real-time updates on passenger presence and other environmental factors.

\textbf{Automated Shuttle:} This is the main vehicle that operates without human intervention. It relies on onboard sensors and external data to navigate and perform tasks.

\textbf{Control Center:} Provides a comprehensive overview of the system. It allows operators to monitor the shuttle's position, status, and interactions with the infrastructure in real time.

In the following subsections, we will outline the architecture of the overall system and provide a detailed description of its components.
\subsection{V2X Communication}
\label{sec:architecture}
The CAM, defined in the EN 302 637-2 standard, is a key message in V2X communication. These periodically broadcast messages contain essential data such as vehicle position, speed, and heading, facilitating safe and efficient traffic management and cooperative driving.
We have developed CAM Messages specifically for the shuttle bus in our extended approach. The shuttle bus position is broadcasted according to its localization, which is further explained in \Cref{sec:shuttle}. Additionally, useful information such as the indicator and door status of the shuttle are incorporated, facilitating tracking of these details in the control center, as outlined in \Cref{sec:control_room}. Encoding of the shuttle mission in CAM Messages allows for simultaneous data recording in the infrastructure, elaborated upon in \Cref{sec:dataset}.
The SPATEM, defined in the TS 103 301 standard, consists of messages from traffic light controllers broadcast by RSUs to share traffic light status with nearby vehicles and road users.
Without real traffic lights, we developed an intelligent traffic light algorithm to publish compatible SPATEM messages via RSUs. To accurately assign the SPATEM messages in the real world, MAPEM road topology information, which is defined in the same standard, was utilized. We manually embedded the MAPEM data for the pedestrian walkway and shuttle driving lane into the RSU. Further applications of this integration are detailed in \Cref{sec:signalization_of_pedestrian_walkway} and \Cref{sec:shuttle}.
The CPM is defined in the TS 103 324 v0.0.22 of the standard. CPMs are standardized messages generated by vehicles and infrastructure to share detailed environmental information with nearby vehicles and other road users.  In this application, the standard implementation of198 198
CPMs highlight their importance in detecting and ensuring the safety of VRUs, enhancing situational awareness, and improving the overall safety of the transportation system, as outlined in \Cref{sec:method:bus_stop}.

\begin{figure}[htbp]
    \centering
    \includegraphics[width=\textwidth]{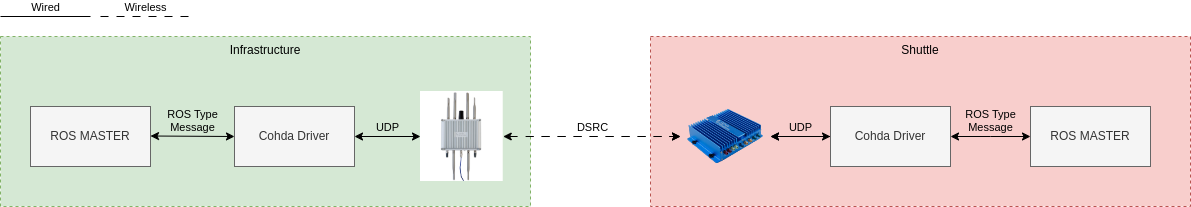}
    \caption{Overview of the information distribution via V2X.}
    \label{fig:driver}
\end{figure}

The standards and our customizations are implemented using our self-developed V2X driver, which will be open-sourced with this paper. This driver connects to V2X COHDA MK6 devices through User Datagram Protocol (UDP) and sends/receives Basic Transport Protocol (BTP) messages. Each message is encoded and decoded using the open-sourced Abstract Syntax Notation One (ASN.1) library\cite{moqvist_asn1tools_2024}. All messages that must be encoded originate from ROS, and after receiving, the messages are decoded and published back into ROS, as illustrated in \Cref{fig:driver}. Each ETSI message type has its own ROS message type to integrate seamlessly into our driving stack.

Importantly, the open-source driver will not be limited to ROS connectivity, ensuring its flexibility and broader applicability. By not restricting it to ROS, users can select their preferred communication framework and develop further accordingly. This Python-based software provides a versatile foundation, empowering a wide range of applications and fostering innovation in V2X communication.

\subsection{Smart Infrastructure}~\label{subsec:smart_infrastructure}%

\label{sec:signalization_of_pedestrian_walkway}
A smart infrastructure implementation is also realized through an intelligent pedestrian crossing. The crossing includes two screens positioned at each pedestrian crossing point, offering clear and intuitive signals through color and symbolic representations, as seen in \Cref{fig:fzi-shuttle:uebergang}. Equipped with V2X technology, the system communicates with the shuttle via SPATEM and CAM messages. SPATEM messages broadcast the current signal phase, allowing the shuttle to adjust its speed accordingly. The system processes CAM messages from the shuttle to dynamically adjust signal phases.
 
The signalization system operates in two modes. The first mode prioritizes shuttle passage when the shuttle is detected within a specific proximity, temporarily halting pedestrian crossing to allow the shuttle to pass safely. The second mode enforces a waiting period for the shuttle, giving pedestrians the opportunity to cross safely before the shuttle proceeds. This mode increases pedestrian flow. When no vehicles are nearby, the system always permits pedestrian crossing, minimizing wait times when it is safe to cross.

The system also estimates the shuttle's arrival time based on its location, direction, and velocity, displaying a countdown on the screens. This allows pedestrians to gauge when the signal phase will change.

This smart signalization system offers several key benefits. It optimizes traffic flow by enabling the shuttle to adjust its approach, reducing unnecessary stops and starts. Additionally, it minimizes pedestrian wait times, enhancing the overall efficiency and safety of the crossing area.

\subsection{Bus Stop}
\label{sec:method:bus_stop}

The proposed intelligent bus stop concept is designed to enhance urban mobility and safety. This system was deployed at two bus stops at the trade fair, each equipped to communicate detected pedestrians to an automated shuttle operating between the stops via V2X technology.

Each bus stop includes several key hardware components. A hemispherical LiDAR sensor with a 90° vertical FOV and 360° horizontal FOV, mounted at the ceiling in the bus stop. It provides a large field of view and captures high-resolution 3D point cloud data of the bus stop and its surroundings. An RSU supports communication and interaction between the bus stop and mobile agents. An integrated compute unit processes the LiDAR data and detects traffic participants locally. Additionally, it manages the communication between the bus stop and agents. All devices in the bus stops are connected via a 2.5 Gigabit ethernet network, also connected to the Control Center, as described in \Cref{sec:control_room}. 
An integrated display presents real-time point cloud data from the LiDAR sensor. 
It visualizes tracked obstacles, providing users with an intuitive understanding of the environment as the bus stop perceives it and assuring pedestrians that they are detected.

\begin{figure}[ht!]
    \centering
    \begin{subfigure}[t]{0.32\linewidth}
        \includegraphics[width=\textwidth]{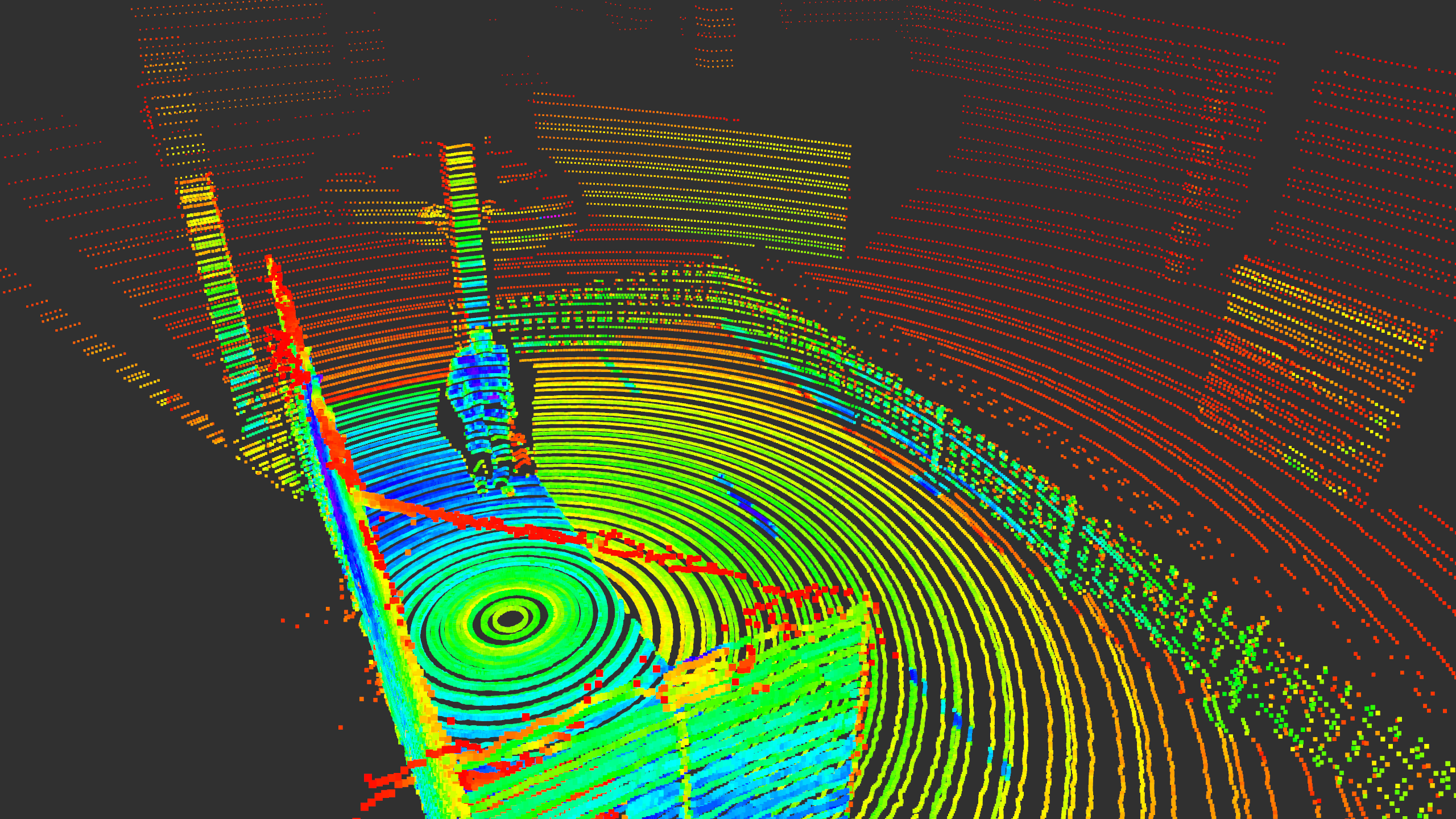}
        \caption{Raw data of the LiDAR sensor with intensity}
        \label{fig:bus-stop:raw}
    \end{subfigure}
    \hfill
    \begin{subfigure}[t]{0.32\linewidth}
        \includegraphics[width=\textwidth]{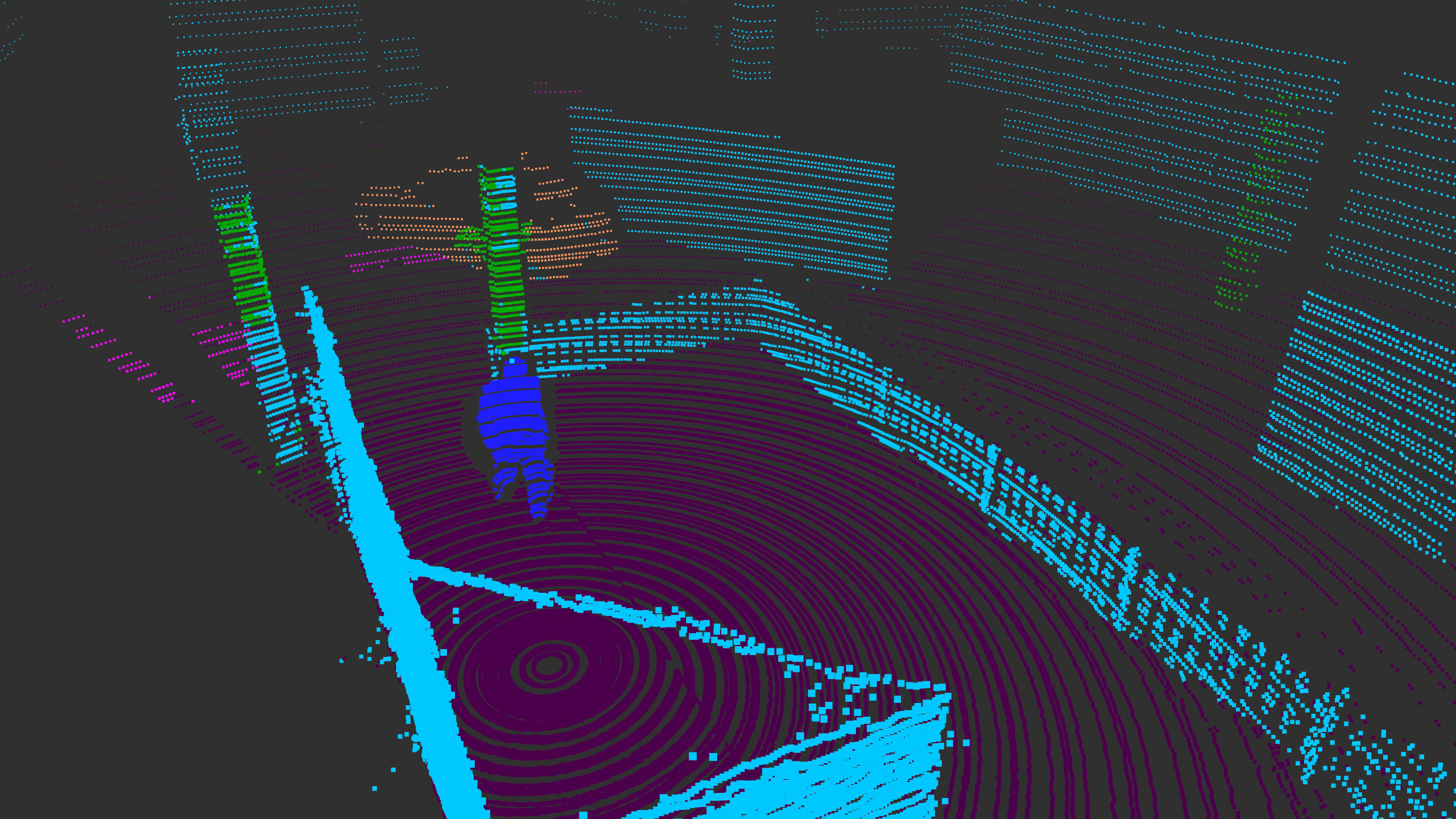}
        \caption{Semantic segmented pointcloud}
        \label{fig:bus-stop:segmented}
    \end{subfigure}
    \hfill
    \begin{subfigure}[t]{0.32\linewidth}
        \includegraphics[width=\textwidth]{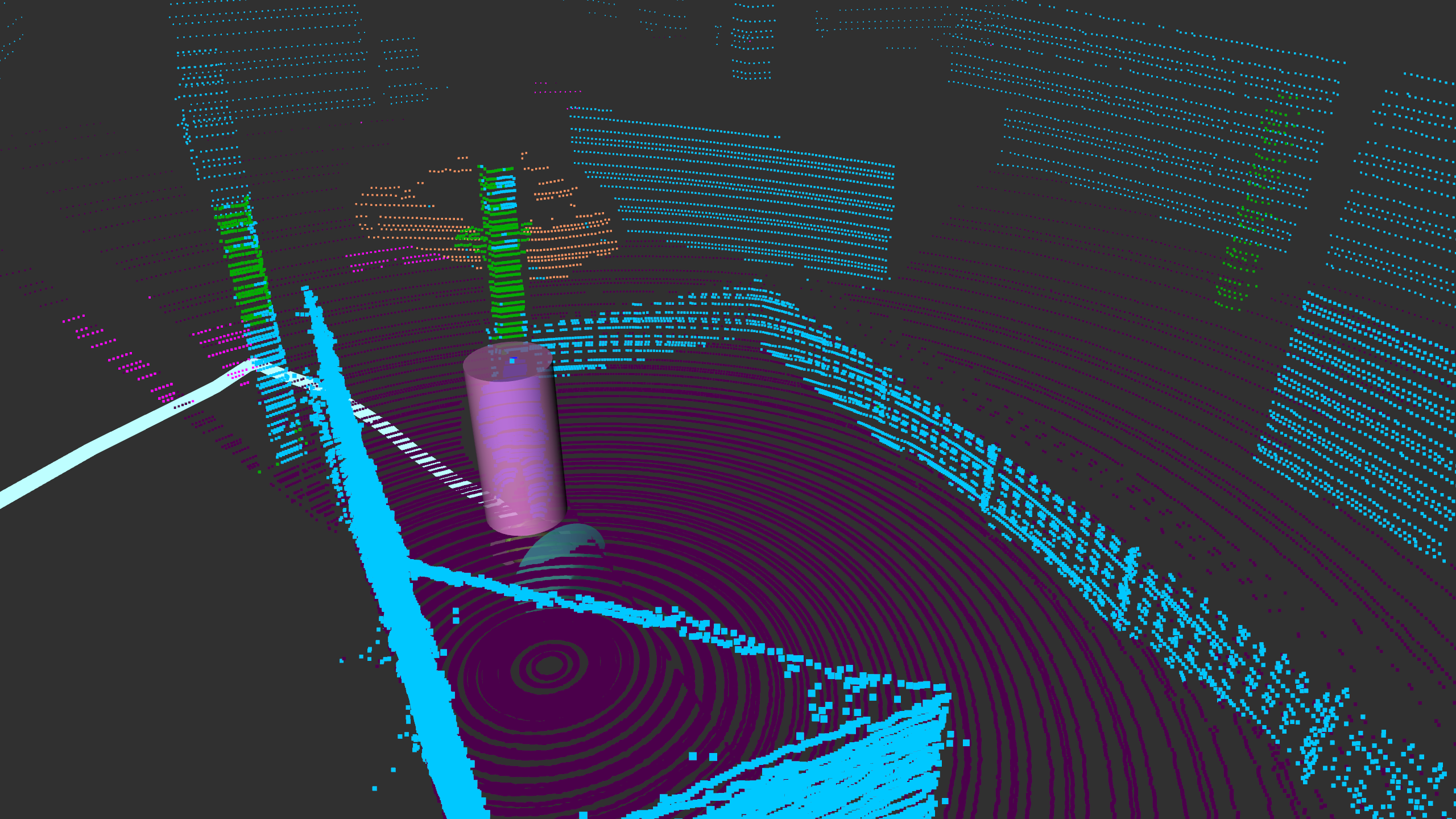}
        \caption{Bounding cylinder with tracked trajectory}
        \label{fig:bus-stop:tracking}
    \end{subfigure}

    \caption{Visualization of the segmented pointcloud with a tracked pedestrian.}
    \label{fig:bus_stop}
\end{figure}

Initially, the local detection pipeline uses a LiDAR segmentation model to classify each point of the LiDAR point cloud. Afterwards a background subtraction module filters out static objects and structures. The remaining segmented non-background points are then clustered and classified objects are predicted. A tracking algorithm assigns a unique ID to each object in order to maintain a consistent identity as they move through the scene. This allows the system to monitor the trajectories and behaviors of traffic participants and predict future movement. An exemplary visualization of the raw pointcloud, the segmented pointcloud and a detected and tracked person are displayed in \Cref{fig:bus_stop}

Both bus stops utilize V2X technology to communicate with V2X-enabled traffic participants. Tracked dynamic obstacles perceived by the bus stop are communicated via CPM messages to increase the perceived region of automated vehicles, improving safety if VRUs are detected on the street. The shuttle also broadcasts its current position and intention with CAMs, allowing the bus stop to display the current position of the shuttle and the estimated time of arrival. The bus stops are also connected via ethernet to the control center and communicate the number of detected pedestrians at the bus stop.

\subsection{Automated Shuttle}
\label{sec:shuttle}
\begin{figure}[ht!]
    \centering
    \begin{subfigure}[t]{0.32\linewidth}
        \includegraphics[width=\textwidth]{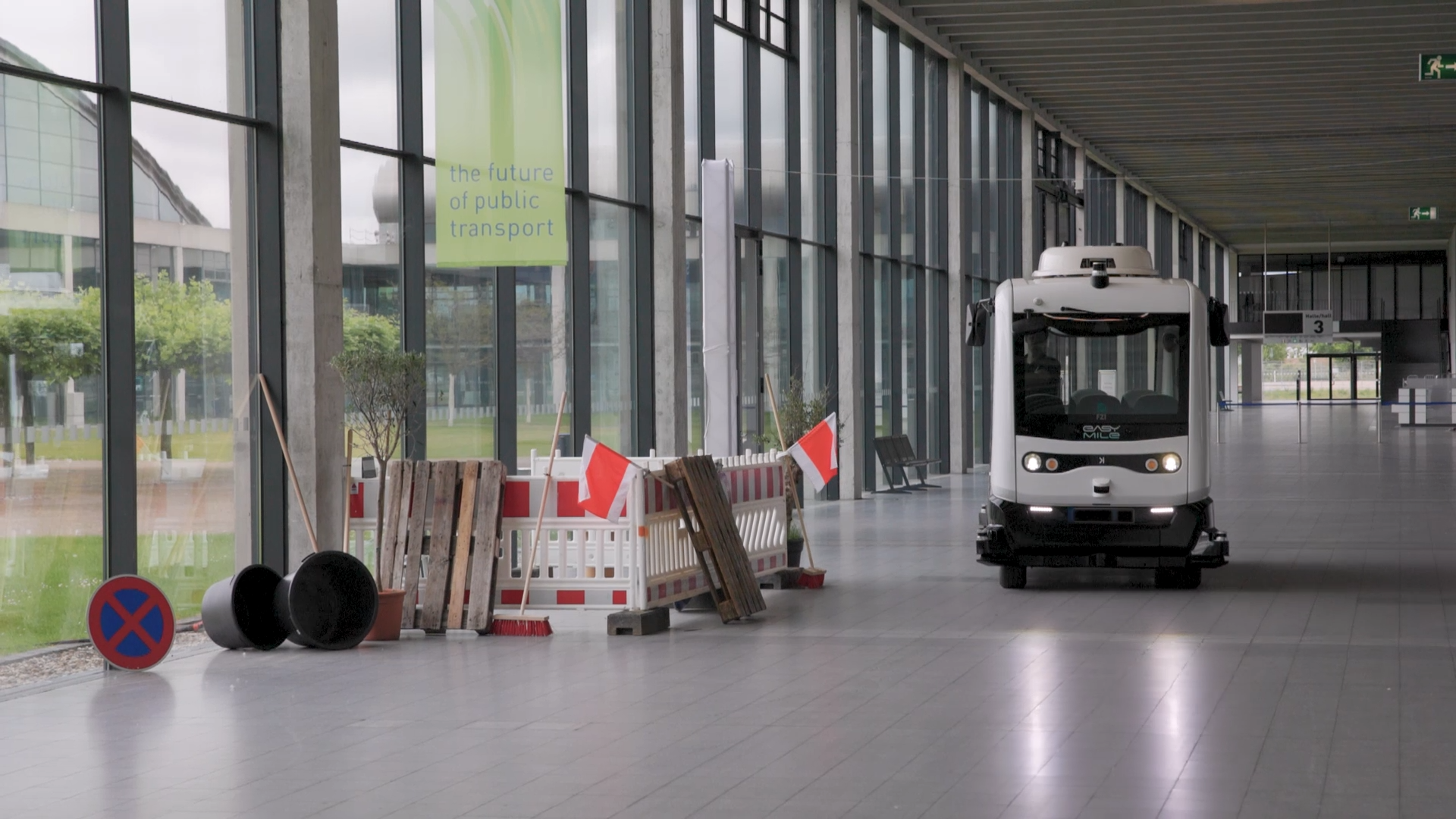}
        \caption{Breaking free from the virtual rail}
        \label{fig:fzi-shuttle:umfahren}
    \end{subfigure}
    \hfill
    \begin{subfigure}[t]{0.32\linewidth}
        \includegraphics[width=\textwidth]{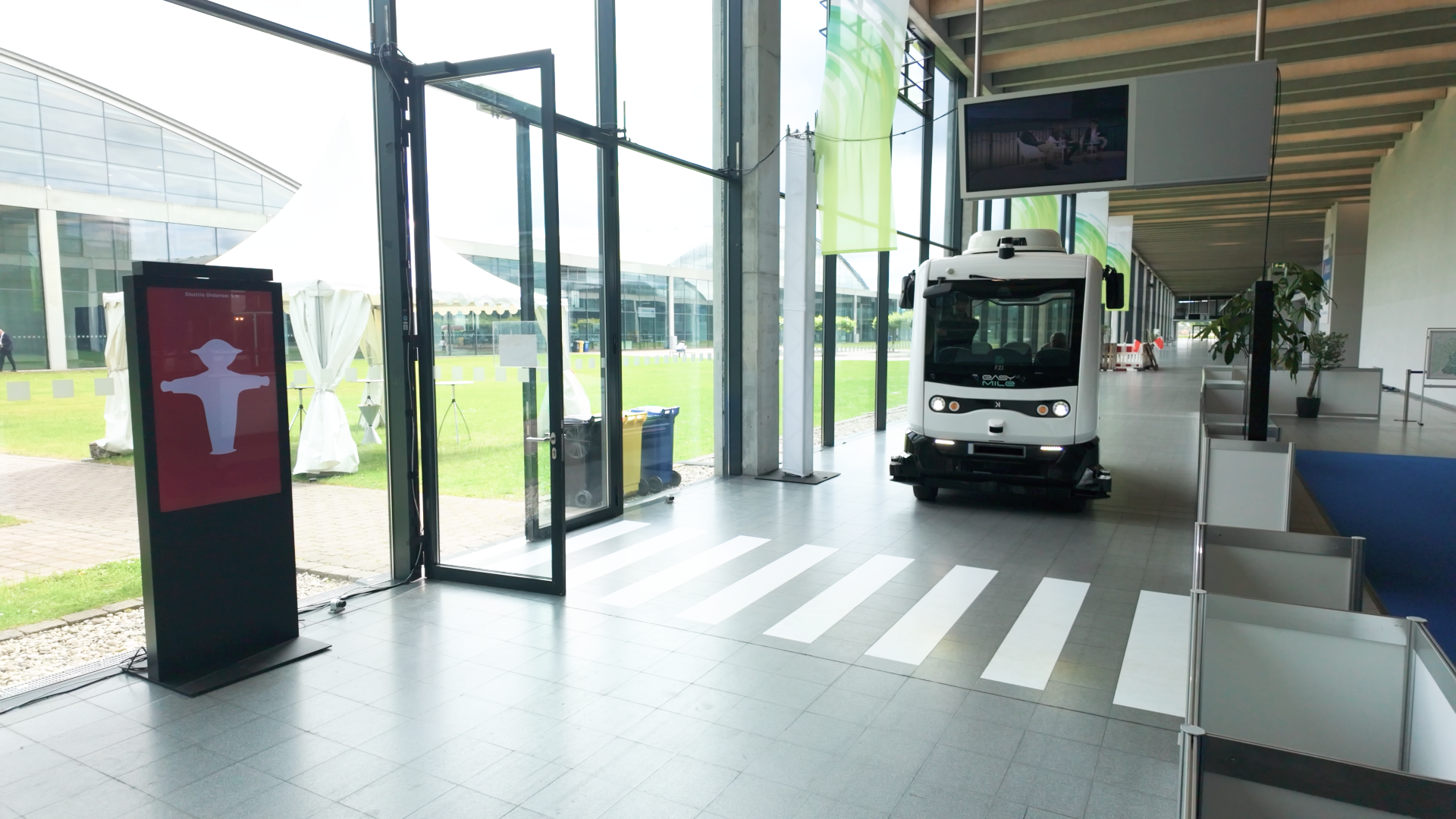}
        \caption{Integration into intelligent transportation systems}
        \label{fig:fzi-shuttle:uebergang}
    \end{subfigure}
    \hfill
    \begin{subfigure}[t]{0.32\linewidth}
        \includegraphics[width=\textwidth]{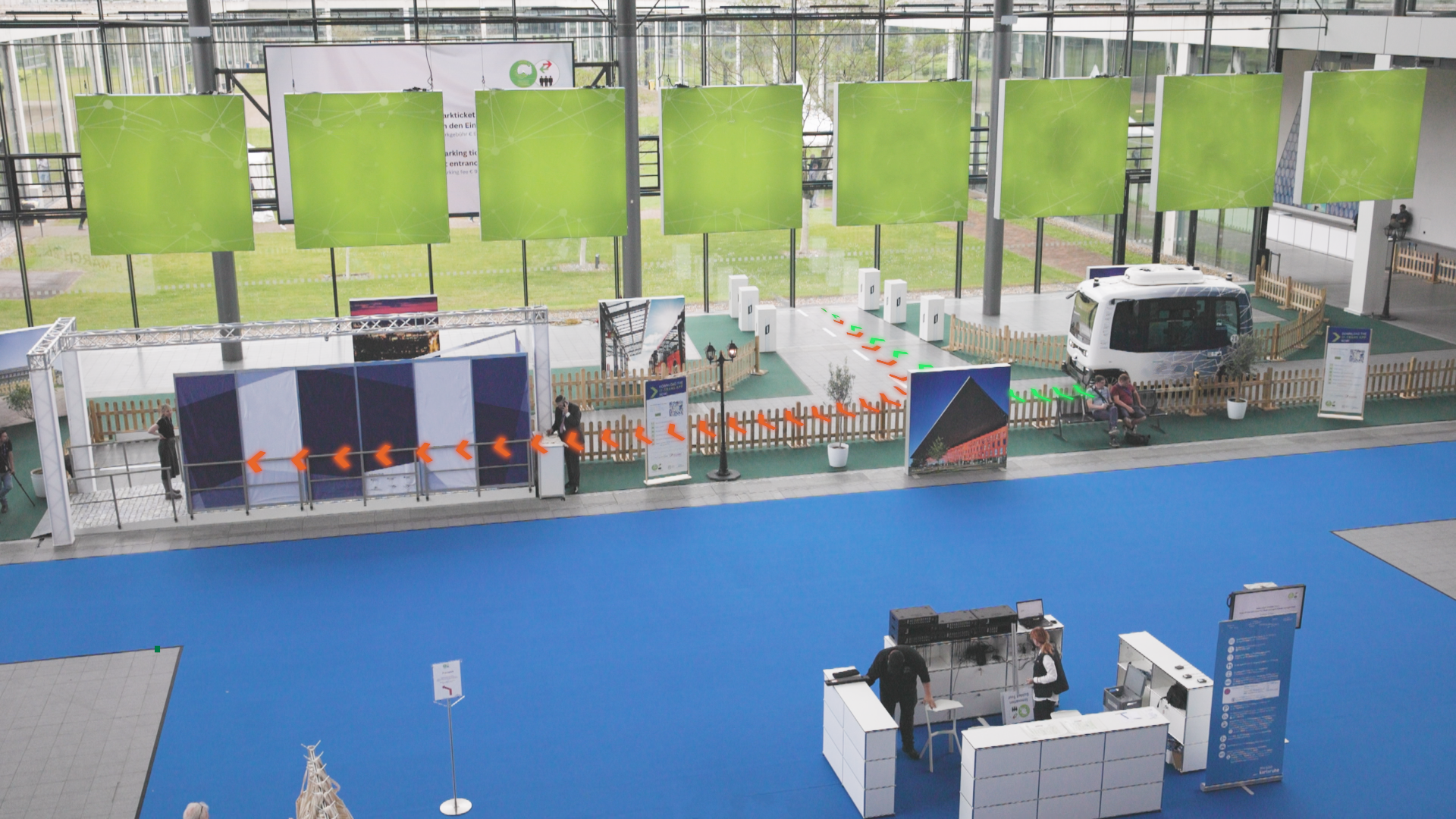}
        \caption{Automtated turning maneuvers}
        \label{fig:fzi-shuttle:maneuver}
    \end{subfigure}
    \caption{Unique selling points of the automated shuttles are the integration into intelligent transportation systems, breaking free from virtual rails and automated turning maneuvers}
    \label{fig:fzi-shuttle}
\end{figure}

The transportation task is fulfilled by an automated shuttle. During \cite{ochs_stepping_2023}, the shuttle was built up and empowered to drive without the constraints of a virtual rail. Breaking free from the concept of a virtual rail allows the shuttles to navigate autonomously around obstacles like in \Cref{fig:fzi-shuttle:umfahren}. These obstacles can be detected from multiple sources. The primary detection of obstacles comes from the onboard sensors using the algorithms presented in \cite{ochs_one_2024}. The shuttle receives additional information from the intelligent infrastructure utilizing CPM and CAM messages. All obstacles are predicted and used in the motion planning module for collision avoidance. Intersection signal phases are received via SPATEM messages and mapped to the high-definition map using the lanelet2 format\cite{poggenhans_lanelet2_2018}. This enables automated yielding at the intersection without error-prone onboard detection of traffic lights. This is shown in the \Cref{fig:fzi-shuttle:uebergang}.

 To handle diverse situations, the shuttle employs multiple planning algorithms. Motion planning based on the particle swarm optimization (PSO) algorithm, as presented in \cite{ochs_leveraging_2024}, is used for driving at high speeds. For sharp and precise turning maneuvers and the positioning at the bus stops, an A*-based~\cite{lavalle_planning_2006} planning approach is used.

The coordination between these algorithms is managed by a broker, internally called switch box. The switch box operates with a hierarchical priority system, always attempting to utilize the highest-priority planning algorithm. When transitioning from one planning algorithm to another, certain constraints must be met: the calculation time must fall within a defined timeframe, and the current position and heading of the active and target trajectories must align within a specified error window.

This validation process allows the switch box to integrate external trajectories from intelligent infrastructure. Consequently, it enhances the ability to teleoperate and navigate the shuttles in tight spaces and congested areas, where the shuttle's internal sensors might only perceive a limited portion of the environment.
\subsection{Control Center}
\label{sec:control_room}

\begin{figure}[htbp]
    \centering
    \includegraphics[width=0.75\textwidth]{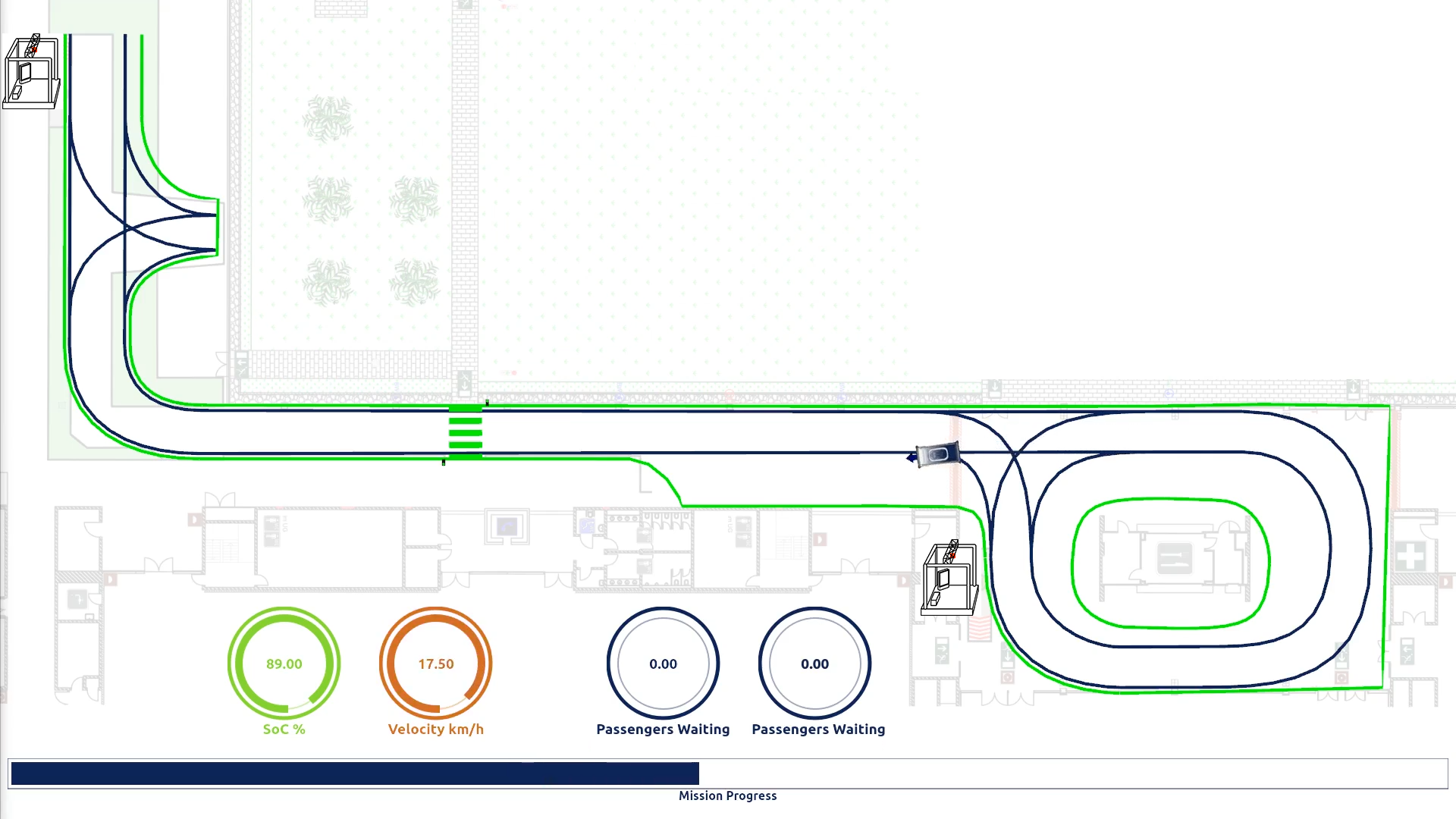}
    \caption{Overview of the collected information of the smart infrastructure, the bus stops, and the automated shuttles.}
    \label{fig:control_center}
\end{figure}

Even in this small-scale experiment, managing the amount of information and diagnosing system failures is challenging. To monitor and supervise all components of the setup, a control center was established. The control center provides a comprehensive overview of the smart infrastructure, the bus stops, and the automated shuttle. This allows the remote supervisor to monitor the operation and, therefore, enhance the safety and efficiency of the operation, as \cite{gontscharow_scalable_2024} showed. 

\Cref{fig:control_center} shows the underlying geometry of the experimental area, including the tracks and driveable area for the automated shuttle. The current position and heading of the shuttle are shown in a top-down view, along with additional dynamic state information such as driving direction and indicator status. The bottom left of \Cref{fig:control_center} features two gauge meters displaying the vehicle's velocity and State of Charge (SoC). This accurate live data allows the remote operator to monitor the shuttle's functionality and ensure safety requirements.

The pedestrian walkway visualizes the traffic light status for pedestrians.
Green indicates safe crossing, while red signals pedestrians to stop.
Passengers waiting at the bus stops are displayed as anonymized bounding boxes in the overview map.
A gauge meter visualizes the number of waiting passengers at each bus stop.
This comprehensive overview allows for effective monitoring, ensuring safe and efficient operation of the shuttle within the defined operational design domain.
Besides the overview map, detailed views of the bus stops with waiting passengers as bounding boxes are also shown on different monitors.

\section{Dataset}
\label{sec:dataset}
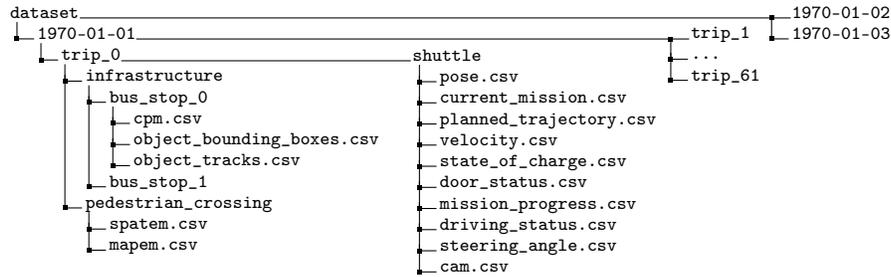
\begin{figure}[htbp]
    \setlength{\DTbaselineskip}{8pt}
    \DTsetlength{0.14em}{0.7em}{0.14em}{0.4pt}{1.6pt}
    \scriptsize
    \centering
    \begin{tikzpicture}
        \def\totalwidth{\textwidth}
        \def\widthminipageone{5.35cm}
        \def\widthminipagetwo{3.66cm}
        \def\widthminipagethree{1.34cm}
        \def\widthminipagefour{1.83cm}
        \node[anchor=north west] (minipage1) at (0, 0) {
            \begin{minipage}[t]{\widthminipageone} %
                \dirtree{%
                    .1 dataset.
                        .2 1970-01-01.
                            .3 trip\_0.
                                .4 infrastructure.
                                    .5 bus\_stop\_0.
                                        .6 cpm.csv.
                                        .6 object\_bounding\_boxes.csv.
                                        .6 object\_tracks.csv.
                                    .5 bus\_stop\_1.
                                .4 pedestrian\_crossing.
                                    .5 spatem.csv.
                                    .5 mapem.csv.
                }
            \end{minipage} %
        };
        \node[anchor=north west] (minipage2) at (\widthminipageone, -0.57cm) {
            \begin{minipage}[t]{\widthminipagetwo} %
                \dirtree{%
                                .1 shuttle.
                                    .2 pose.csv.
                                    .2 current\_mission.csv.
                                    .2 planned\_trajectory.csv.
                                    .2 velocity.csv.
                                    .2 state\_of\_charge.csv.
                                    .2 door\_status.csv.
                                    .2 mission\_progress.csv.
                                    .2 driving\_status.csv.
                                    .2 steering\_angle.csv.
                                    .2 cam.csv.
                }
            \end{minipage} %
        };
        \node[anchor=north west] (minipage3) at (\widthminipageone+\widthminipagetwo, -0.2) {
            \begin{minipage}[t]{\widthminipagethree} %
                \vspace{2pt}
                    \dirtree{%
                                .1 \,trip\_1.
                                .1 \dots.
                                .1 trip\_61.
                    }
            \end{minipage} %
        };
        \node[anchor=north west] (minipage4) at (\widthminipageone+\widthminipagetwo+\widthminipagethree, 0) {
            \begin{minipage}[t]{\widthminipagefour} %
                \dirtree{%
                        .1 \,1970-01-02.
                        .1 1970-01-03.
                }
            \end{minipage} %
        };
    \draw[black, line width=0.4pt] (1.85, -0.87) -- (5.7, -0.87); %
    \draw[black, line width=0.4pt] (2.05, -0.58) -- (9.4, -0.58); %
    \draw[black, line width=0.4pt] (1.3, -0.3) -- (10.75, -0.3); %

    \fill[white] ([shift={(-0.97em,-0.4pt)}]9.4,-0.58) rectangle ([shift={(1.6pt-0.97em,3pt)}]9.4,-0.58); %
    \fill ([shift={(-0.97em,-0.8pt)}]9.4,-0.58) rectangle ([shift={(1.6pt-0.97em,0.8pt)}]9.4,-0.58); %
    \fill[white] ([shift={(-1em,-0.4pt)}]10.75,-0.3) rectangle ([shift={(1.6pt-1em,3pt)}]10.75,-0.3); %
    \fill ([shift={(-1em,-0.8pt)}]10.75, -0.3) rectangle ([shift={(1.6pt-1em,0.8pt)}]10.75, -0.3); %

\end{tikzpicture}
    \caption{Structure of the dataset.}
    \label{fig:folder_structure}
\end{figure}
We provide a dataset in form of CSV-File recordings with the publication of this paper. The dataset was recorded during a three-day trade fair, where the setup is demonstrated in~\cref{sec:method}. It contains 207~individual recordings, each recording containing one trip between the two bus stops, with a total driving duration of~7~hours and~28~minutes. 
\cref{fig:folder_structure}~shows the structure of the published dataset\footnote{The dates aren't set appropriately due to the double-blind review process and will be set after paper acceptance.}. 
The recorded data can be divided into three sources: data from the smart intersection (see \Cref{sec:signalization_of_pedestrian_walkway}), data from the smart bus stops (see \Cref{sec:method:bus_stop}), and data from the shuttle itself (see \Cref{sec:shuttle}). Data from the two smart bus stops and the smart intersection was recorded as a rosbag in the infrastructure. Similarly, data from the shuttle was recorded as a rosbag but stored on the shuttle itself. These rosbags were merged afterwards based on time stamps. Recording in the shuttle was triggered by the safety driver selecting a mission. The start of the mission was broadcasted by the shuttle via CAM messages, prompting the infrastructure to start recording as well. 
The bus stop data contains tracked bounding boxes of the pedestrians from the LiDAR data for both bus stops and the corresponding CPM messages. 
From the shuttle, the pose, including position and heading, as well as the current mission and planned trajectory, is included. Additional information, such as the velocity, state of charge, door status, mission progress, driving status (autonomous or manual), and steering angle is also available. The CAM messages from the shuttle are also included. 
The data from the pedestrian walkway includes SPATEM and MAPEM messages, that is, the current traffic light phase and the time until the next phase change and the geometry information of the pedestrian crossing.
\section{Evaluation}
\label{sec:evaluation}
\subsection{Package Loss Evaluation }
\label{sec:evaluation:package_loss}
\begin{figure}[ht!]
    \centering
    \begin{subfigure}[t]{0.48\linewidth}
        \includegraphics[width=\textwidth]{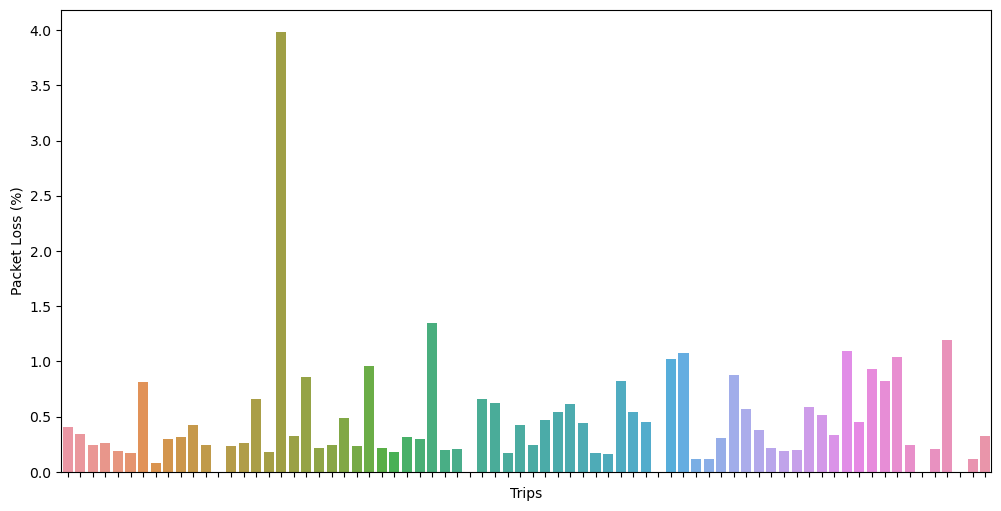}
        \caption{Analysis of package loss during one day.}
        \label{fig:packageloss}
    \end{subfigure}
    \begin{subfigure}[t]{0.48\linewidth}
    \centering
        \includegraphics[width=\textwidth]{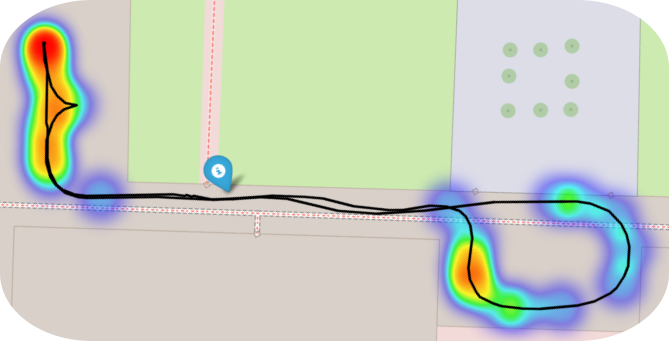}
        \caption{Heatmap of package loss during one day. Visualization is based on OpenStreetMap\cite{openstreetmap-contributors_planet_2017}}
        \label{fig:packageloss_trip_heatmap}
    \end{subfigure}
    \caption{Analysis of package loss}
    \label{fig:eval_packageloss}

\end{figure}

    \paragraph{}
    As described in the~\Cref{sec:architecture}, the entire system can communicate through the V2X messages. Here, the CPM package loss throughout the route is evaluated. The~\Cref{fig:packageloss} displays a bar chart that quantifies the percentage of lost messages for various recorded trips. Each bar represents an individual trip, while the y-axis indicates the corresponding package loss rate. The~\Cref{fig:packageloss_trip_heatmap} integrates a heatmap on the route to represent the spatial distribution of package losses visually.
    In~\Cref{sec:method:bus_stop}, the bus stops are described. At bus stop 1, signal reliability faltered due to glass windows between the shuttle and the RSU at the pedestrian walkway, resulting in higher package loss. This is notable as these areas typically have more messages due to lower vehicle velocity and turning maneuvers. In contrast, bus stop 2 experienced less package loss despite physical obstructions like a concrete block, indicating varied environmental impacts on V2X message transmission. These temporal and spatial analyses provide a comprehensive perspective on package loss patterns, which is crucial for developing resilient vehicular communication systems to minimize message loss and enhance overall system reliability.

\subsection{Impact of Smart Intersection}

As described in \Cref{sec:signalization_of_pedestrian_walkway}, the system allowed pedestrians to cross the lane whenever the automated shuttle was not near the intersection. Over the course of three days, the setup was active for a total of 19 hours and 56 minutes. During this period, the red light for pedestrians was displayed for only~63~minutes and~33~seconds. Consequently, pedestrians were allowed to cross immediately~\SI{94.6}{\percent} of the time.

\subsection{Evaluation of travel time}
\begin{figure}[ht!]
    \centering
    \begin{subfigure}[t]{0.58\linewidth}
        \includegraphics[height=5cm]{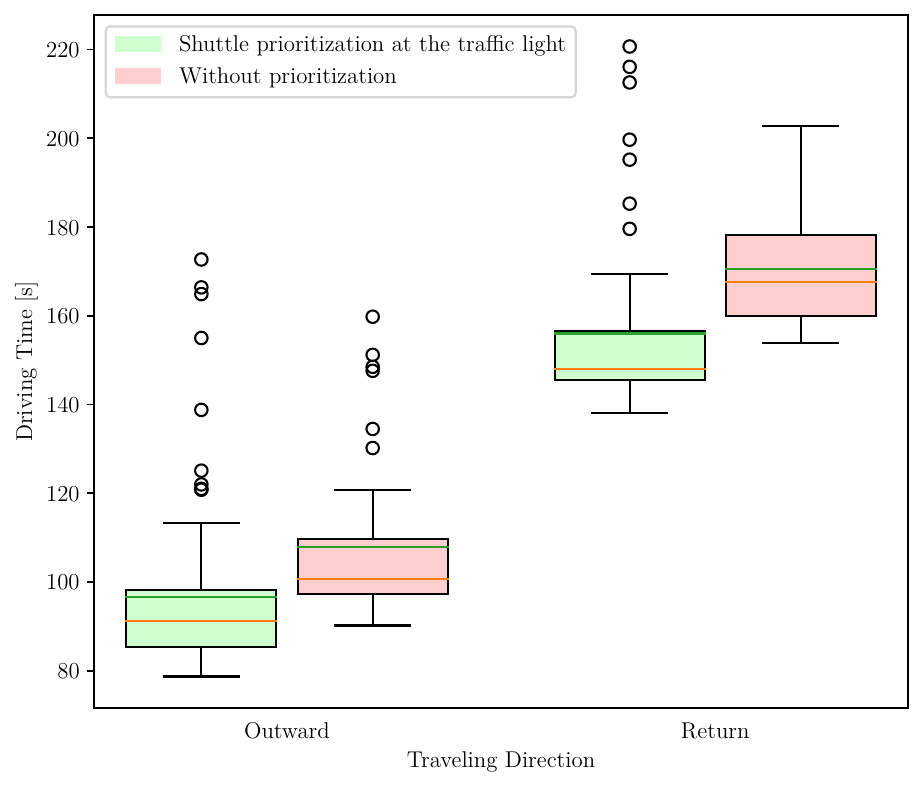}
        \centering
        \caption{Traveling times of the shuttle grouped by driving direction and traffic light mode. The green boxes represent the journeys where the shuttle was prioritized at the pedestrian walkway, the red ones represent the journeys where the shuttle wasn’t prioritized. }
        \label{fig:traveling_times}
    \end{subfigure}
    \hfill
    \begin{subfigure}[t]{0.35\linewidth}
        \includegraphics[height=5cm]{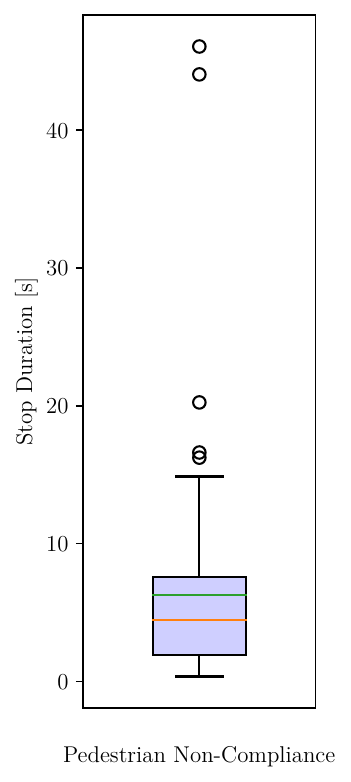}
        \centering
        \caption{Analysis of stop times due to pedestrian non-compliance at smart intersections.}
        \label{fig:non_compliance}
    \end{subfigure}
    \caption{Evaluation of traveling times and effects of pedestrian non-compliance to traffic signs. The median is marked orange and the mean is marked green}
    \label{fig:eval_graphs1}
\end{figure}

\Cref{fig:traveling_times}~shows the traveling time from one bus stop to the other. The average driving time over all journeys of the three-day trade fair is~\SI{124}{\second}, the median travel time is~\SI{109}{\second} with a standard deviation of~\SI{35}{\second}. As shown in~\Cref{fig:traveling_times} the journeys can be divided in two directions, outbound (from bus stop~1 to bus stop~2) and return (from bus stop~2 to bus stop~1). Furthermore, a distinction can be made as to whether the shuttle is prioritized at the pedestrian crossing and can pass directly, or whether it has to stop. These two modes are described more detailed in~\Cref{sec:signalization_of_pedestrian_walkway}. The prioritized journeys are shown in green, the others in red. In each box of the boxplot, the orange line marks the median, while the green line marks the mean. It is easy to see that the travel time is significantly longer in the return direction, which can be attributed to the fact that a turning maneuver has to be carried out during this journey. It can also be seen that the travel time is shortened when the shuttle is prioritized at the pedestrian crossing. The outliers in the travel time can be explained by non-compliant behavior of pedestrians at the pedestrian crossing (for further evaluation on this effect see~\Cref{sec:evaluation:pedestrian_non_compliance}).

\subsection{Impact of Pedestrian Non-Compliance}
\label{sec:evaluation:pedestrian_non_compliance}
In our live demonstration of automated shuttle operations at a trade fair, we observed frequent instances of pedestrian non-compliance with traffic signals. Out of 208 intersection crossings analyzed, pedestrians crossed against red signals intended for them 65 times. We observed that, in most instances, pedestrians knowingly tested the capabilities of the automated shuttle. This behavior forced the shuttle to halt unnecessarily despite having the right of way. The delay, displayed in \Cref{fig:non_compliance} caused by such incidents exceeded 40 seconds in some cases.

This issue arises from the expectation that autonomous vehicles will always yield to pedestrians regardless of traffic signals. This perception can lead pedestrians to ignore red lights, thus compromising operational efficiency. Furthermore, this behavior poses safety risks as autonomous vehicles may not anticipate or react swiftly to unexpected pedestrian movements. Failure to address this problem could undermine the benefits of smart intersections in improving traffic flow and safety in urban environments.

\section{CONCLUSION}
\label{sec:conclusion}

The paper presents a unique integration of several vital components in autonomous transportation with infrastructure support. In this way, the next generation's bus stops can actively support the shuttles from signalizing a demand for transportation to the expansion of the sensor of the automated shuttle. This experimental setup was built indoors with an additional smart infrastructure consisting of an intelligent pedestrian crossing. All these components are conntected via different ETSI standards, like CAM, CPM, SPATEM, and MAPEM.
To conclude the setup, a control center was built to visualize all the dynamic information. Each component had a challenge itself. The bus stops utilizes state-of-the-art semantic segmentation on LiDAR data to detect people. The smart infrastructure had to cope with intelligent traffic flow management and the broadcast range of the RSU in the indoor environment. Last, the automated shuttle had to conduct turning maneuvers to approach the bus stops precisely. All these components were tested in a small-scale operation and demonstrated their potential for scalability, particularly in the control center. However, several areas still require improvement: the shuttles need to operate without a safety operator to achieve true scalability, the detection of passengers at bus stops needs enhancement, and the smart infrastructure must be capable of handling interactions with non-automated vehicles.

\section*{ACKNOWLEDGMENT}

This work was partially funded by KAMO: Karlsruhe Mobility High Performance Center (www.kamo.one).
\bibliographystyle{splncs04}
\bibliography{references}
\end{document}